\documentclass[letterpaper, 10 pt, conference]{ieeeconf}  

\IEEEoverridecommandlockouts                              

\usepackage{color}
\usepackage{caption}
\usepackage{graphics} 
\usepackage{amsmath} 
\usepackage{amssymb}  
\usepackage{biblatex} 
\usepackage{booktabs}
\usepackage{graphicx}
\usepackage[capitalise]{cleveref}

\title{\LARGE \bf
Overcoming Obstructions via Bandwidth-Limited \\Multi-Agent Spatial Handshaking
}

\author{Nathaniel Glaser, Yen-Cheng Liu, Junjiao Tian, and Zsolt Kira\\
Georgia Institute of Technology\\
\texttt{$\{$nglaser,ycliu,jtian73,zkira$\}$@gatech.edu}
}
\begin{document}

\maketitle
\thispagestyle{empty}
\pagestyle{empty}

\begin{abstract}
In this paper, we address \textit{bandwidth-limited} and \textit{obstruction-prone} collaborative perception, specifically in the context of multi-agent semantic segmentation.  This setting presents several key challenges, including processing and exchanging unregistered robotic swarm imagery.
To be successful, solutions must effectively leverage multiple non-static and intermittently-overlapping RGB perspectives, while heeding bandwidth constraints and overcoming unwanted foreground obstructions.  As such, we propose an end-to-end learn-able Multi-Agent Spatial Handshaking network (MASH) to process, compress, and propagate visual information across a robotic swarm.  Our distributed communication module operates directly (and exclusively) on raw image data, without additional input requirements such as pose, depth, or warping data.  We demonstrate superior performance of our model compared against several baselines in a photo-realistic multi-robot AirSim environment, especially in the presence of image occlusions.  Our method achieves an absolute $11\%$ IoU improvement over strong baselines.
\end{abstract}

\section{INTRODUCTION}
Deep learning has revolutionized robotic perception, especially in areas such as object detection and semantic segmentation. 
As a result of these improvements, we are increasingly seeing these algorithms applied at scale, as with self-driving cars and autonomous delivery drones. 
However, many of these core algorithms are developed around \textit{single} agent perception.  
In a world where physical hardware, compute power, and inter-connectivity are becoming increasingly cheap, significant advantages exist in developing these algorithms towards distributed multi-agent systems.

One such advantage is that of information \textit{redundancy}--even if a single robot faces a challenging input or failure mode (such as a local image occlusion), it can leverage the information of its robot collaborators to overcome it.  However, several practical challenges remain: relevant information must be identified, communication bandwidth constraints must be respected, and computation must be effectively distributed across the networked robots.

In this paper, we address these challenges with a learned architecture that identifies and exchanges information across several robots.  Our problem formulation and solution draw inspiration from the learning-based works of several domains, including learned cost volume image correspondence~\cite{melekhov2019dgc,fischer2015flownet,ilg2017flownet,xu2017accurate,wang2018occlusion}, learned communication~\cite{liu2020who2com,liu2020when2com}, and learned multi-view fusion~\cite{su2015multi,multiviewinpainting2016,rgbdInpainting2018}.  Based on these prior works, we develop a novel architecture that effectively combines information from multiple robots in a distributed, bandwidth-limited manner, without assuming known positions or transformations between their sensors.  

Unlike prior work, we introduce several novel features that allow for a tune-able improvement in perception accuracy, as dictated by bandwidth constraints.  We extend the 1D handshaking mechanism of When2Com~\cite{liu2020when2com} to a 2D spatial handshaking mechanism, allowing for patch-based identification, communication, and exchange of visual information across agents.  As with prior work, we match image patches via cost volume correspondence; however, in contrast to previous methods, we address \textit{multi-agent} correspondence and introduce a novel correspondence autoencoder, which infills the spurious matches and obstructed patches that are common in an occlusion-prone, multi-agent setting.  Finally, we duplicate our cross-agent patch-matching pipeline across several agents and perform multi-agent fusion on warped and exchanged segmentation data.

\begin{figure}
\vspace{2mm}
\centering
\includegraphics[width=\linewidth]{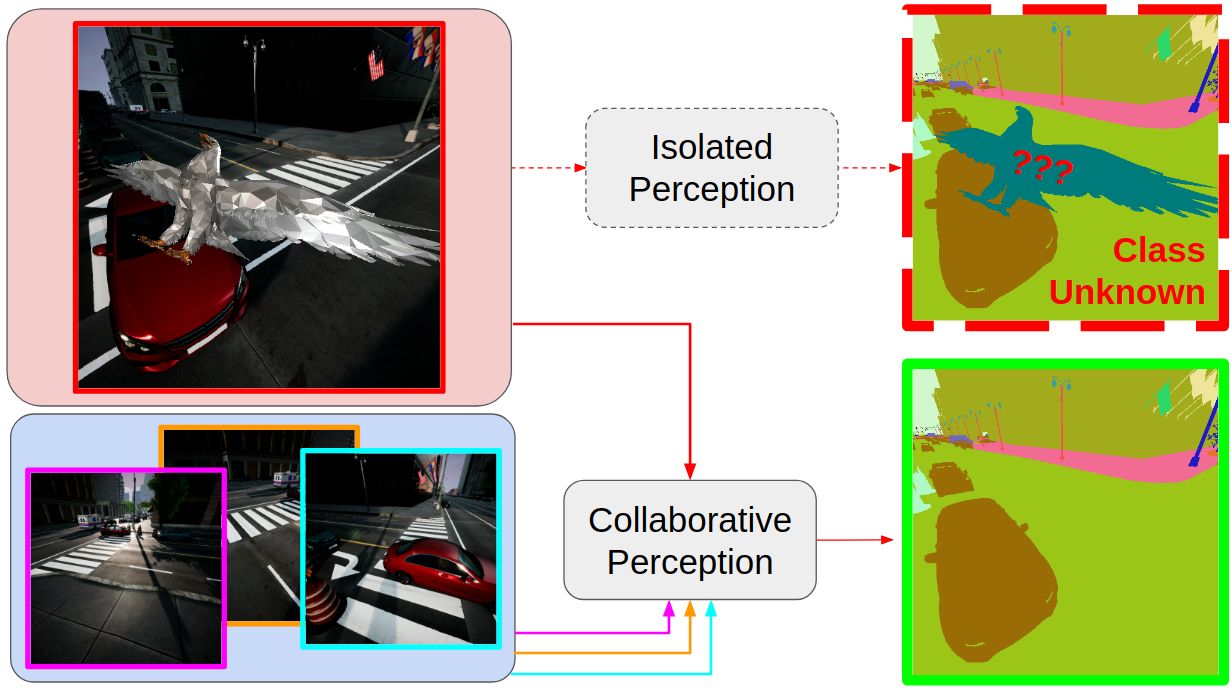}
\caption{\textbf{Collaborative Perception}.  Alone, an isolated perception system may produce an undesirable output in the presence of an unexpected obstruction.  On the other hand, a collaborative perception system may leverage multiple views to overcome potentially dangerous misclassifications.}
\label{fig:Qualitative_Infill}%
\end{figure}

To demonstrate the efficacy of our approach, we perform extensive experimentation in a photo-realistic simulation environment (AirSim~\cite{airsim2017fsr}), specifically investigating how a group of mobile robots can communicate their observations to overcome unexpected foreground obstructions, such as occluding vegetation and wildlife. Our method increases performance on a multi-agent semantic segmentation task by an absolute $11\%$ IoU over strong baselines, and it approaches upper bounds that utilize ground truth transformations across the agents' sensors, while saving significant bandwidth.

We summarize the contributions of our paper as follows:
\begin{itemize}
    \item We introduce a variant on the \textit{Multi-Robot Collaborative Perception} task~\cite{liu2020when2com, liu2020who2com} in which a group of independent robotic agents captures synchronized 2D RGB data as they travel through a shared 3D environment.
    Unlike prior work, we do not provide any pose and depth information during inference.
    \item To address this challenge, we propose an end-to-end learn-able \textbf{M}ulti-\textbf{A}gent \textbf{S}patial \textbf{H}andshaking network, \textbf{MASH}. We contribute a set of novel techniques that address challenges that the current state of art do not address, including extending distributed communication mechanisms into 3D, introducing a smoothing autoencoder to address potential inaccuracies in cost volume estimation, and adding a tune-able compression mechanism for trading-off bandwidth and perception accuracy. 
    \item We test our algorithm on the multi-robot collaborative perception task within a photorealistic simulator, demonstrating superior performance. We specifically highlight a real-world application of collaborative perception with an \textit{obstruction-prone} multi-drone dataset.  
\end{itemize}
\section{RELATED WORK}
Our work builds on several domains, including image correspondence, optical flow, visual odometry, visual SLAM, communication, inpainting, and multi-view fusion.  Here, we highlight the learning-based approaches from each domain and describe how each contributes to the design of our network architecture. 

\textbf{Image Correspondence} methods explore how to correspond visual information between pairs of images.  To perform this correspondence, learning-based approaches typically use learned feature extraction and feature comparison modules.  Prior work ranges from sparse matching via learned interest points~\cite{detone2017toward,detone2018superpoint} to dense matching with patches~\cite{rocco2017convolutional} and cost volumes~\cite{melekhov2019dgc}.  These techniques are particularly relevant to our work, which seeks to correspond spatial image information between pairs of independent agents.  Specifically, we extend the cost volume decoder of DGC~\cite{melekhov2019dgc} into a cost volume \textit{autoencoder}.  We train the autoencoder to use spatial \textit{context} and matching \textit{uncertainty} to infill a warping grid and produce warping scores between pairs of agents.

\textbf{Visual SLAM} corresponds image features across observations but also creates a spatial representation (\textit{i.e.} a map) of these features.  This spatial representation allows for the features to be more reliably corresponded and referenced for localization estimates.  A notable learning-based example of visual SLAM is EMPNet~\cite{avraham2019empnet} which uses a dense depth map to reproject image patches (and associated learned feature embeddings) into a 3D map.  Subsequent 3D reprojections are then corresponded via weighted point cloud alignment to build the map and to produce localization estimates.  Though inspired by the learned correspondence of visual SLAM, our work addresses a different problem setting.  Rather than building an accurate map from a continuous sequence of observations collected over time, we instead focus on how to best leverage currently available sensor information for an immediate prediction task, especially from several independent and disparate viewpoints.  

\textbf{Learned Communication} methods exchange information across multiple observers via learned features.  Who2Com~\cite{liu2020who2com} and When2Com~\cite{liu2020when2com} use a learned handshake communication protocol to pass complementary information between degraded and non-degraded agents.  Specifically, each agent broadcasts a learned \textit{query} vector to the others; the other agents compare this \textit{query} against an internal \textit{key} vector to produce a matching score; this matching score is used to forward and fuse \textit{feature} vectors from contributing agents for eventual decoding into the final task output.  Our work extends this one-dimensional handshaking mechanism into a two-dimensional \textit{spatial} handshaking mechanism.  This extension allows the network to efficiently flag patches of an image for bandwidth-limited exchange across agents.

\textbf{Inpainting} methods attempt to replace missing or corrupted regions of an image based on surrounding image context.  The blind inpainting convolutional neural network (BICNN)~\cite{cai2017blind} overcomes image corruption by learning from a dataset that contains corrupted images and their non-corrupted counterparts.  The work of Purkait et al~\cite{purkait2019seeing} employs a similar method to infill the occluded semantic masks of foreground and background objects.  Similarly, our work addresses the same task of overcoming image occlusions and degradations.  However, unlike inpainting works which ``hallucinate'' plausible replacements for occlusions, our work leverages the perspectives of multiple agents to ``see past'' those occlusions.

\textbf{Multi-View Fusion} combines multiple views into a shared representation that can then be decoded for some task.  For instance, the Multi-View Convolutional Neural Network (MVCNN)~\cite{su2015multi} classifies 3D objects by combining the features of multiple views into a single representation with a view-pooling operation.  Non-learned multi-view fusion, as with multi-view inpainting~\cite{multiviewinpainting2016, rgbdInpainting2018}, use 3D reconstruction techniques or depth data to warp image data between viewpoints; additionally, these works tackle the issue of missing correspondences via patch-based or exemplar-based infilling.
As with these prior works, our work fuses information from multiple views.  However, unlike MVCNN and other learned multi-view methods, our approach explicitly learns how to align spatial images features, rather than implicitly learning these transformations as part of the network weights.  This explicit alignment allows for better interpretability and more flexibility with the features selected for cross-agent exchange and fusion, as shown by our experimental results.  Furthermore, unlike many of these multi-view works, our approach does not require depth data, relative pose data, static viewpoints, or expensive 3D reconstruction techniques.  Without these restrictions, our method is sufficiently lightweight for distributed deployment across multiple mobile agents.
\section{METHOD}

\begin{figure*}
\vspace{5mm}
\centering
\includegraphics[width=\linewidth]{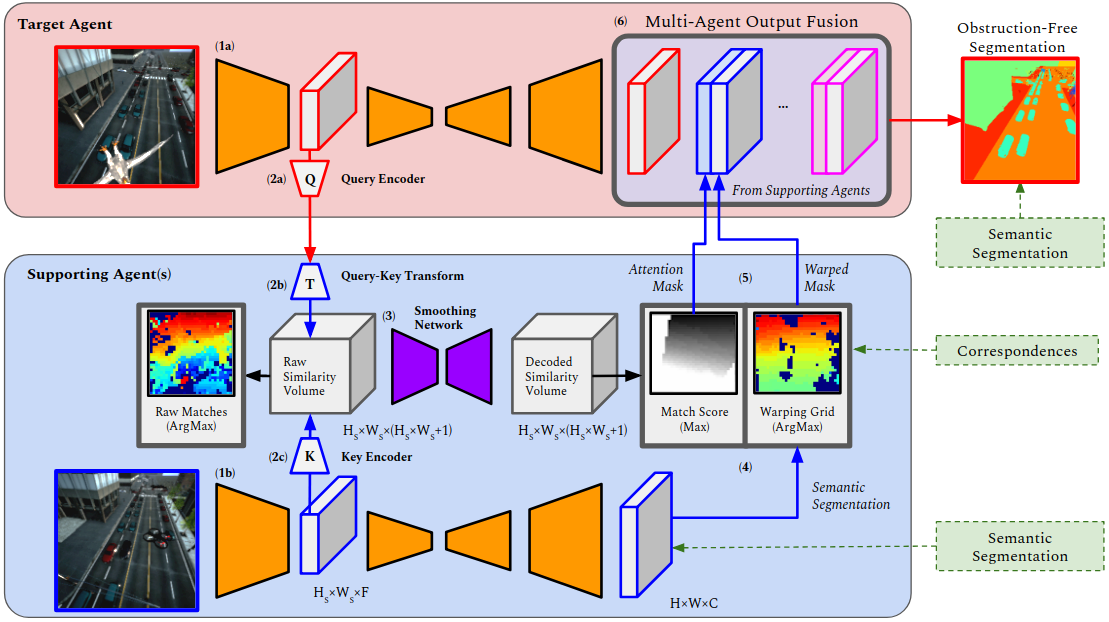}
\setlength{\belowcaptionskip}{-15pt}
\caption{\textbf{Multi-Agent Spatial Handshake Network}: This figure depicts the spatial handshake between a \textbf{target} agent (red block) and one of its \textbf{supporting} agents (blue block).  
Without loss of generality, we describe the spatial handshaking procedure for one pair of agents: (1ab) a target agent and a set of supporting agents use a segmentation network (orange trapezoids) to compute spatial feature maps and a semantic segmentation mask; (2a) the target agent compresses its feature map with a \textbf{query encoder} and broadcasts it to all supporting agents; (2b) each supporting agent receives this compressed map and reshapes it with a \textbf{query-key transform} to handle any asymmetry between the query and key; (2c) each then compares this reshaped map with its own feature map similarly compressed by a \textbf{key encoder}; (3) the resulting similarity volume is smoothed with an autoencoder \textbf{smoothing network} (purple trapezoids); (4) channel-wise \textbf{Max} and \textbf{ArgMax} operators then generate a 2D matching score and warping grid, respectively; (5) each supporting agent uses its warping grid to sample its output distribution; and (6) the corresponding match scores are used as the spatial attention weights during \textbf{multi-agent fusion}.  The green boxes highlight the ground truth information used to train the network.}
\label{fig:overall}%
\end{figure*} 

\subsection{Multi-Agent Spatial Handshaking}\label{sec:MASH}
The \textit{Multi-Agent Collaborative Perception} task involves producing the most accurate \textit{egocentric} semantic segmentation prediction for one robot (or set of robots), given bandwidth-limited access to the current observations of its collaborators.  To achieve this goal, each robot must identify complementary visual information despite \textit{intermittently} and \textit{partially} overlapping fields of view with the others.  Additionally, these robots must be able to correspond their observations in the presence of unknown obstructions (\textit{i.e.} close-up objects that are distinct from the known classes).  Together, these components probe one of the challenges that faces a networked autonomous vehicles: namely, how can we \textit{efficiently} leverage peers to improve perception performance and resistance to visual occlusions and outliers?

The data associated with this task includes high-resolution inputs (RGB images), ground truth outputs (semantic segmentation masks), and auxiliary training information (pairwise pixel correspondences).  Additionally, communication bandwidth limitations prevent direct transmission of the raw input data; therefore, information must be efficiently processed and exchanged.

To this end, we propose an extensible neural pipeline that leverages several components inspired by prior work.  Our overall system uses segmentation networks~\cite{badrinarayanan2017segnet} to extract features, handshake communication~\cite{liu2020when2com} to compress these features for transmission, cost volume correspondences~\cite{fischer2015flownet,ilg2017flownet,xu2017accurate,detone2016deep,wang2017deepvo} to associate them, and attention mechanisms~\cite{vaswani2017attention} to fuse the associated data.  In order to solve the challenging collaborative perception problem with unknown correspondences, we introduce three significant novel aspects: (1) we extend the 1D \textbf{key}-\textbf{query}-\textbf{feature} handshaking mechanism of When2Com~\cite{liu2020when2com} into a 2D spatial handshaking mechanism for data compression and transmission; (2) we introduce a novel smoothing autoencoder that smooths spurious matches and infills obstructed matching regions; and (3) we duplicate our dense correspondence pipeline across several agents and perform multi-agent fusion on the warped and scored segmentation results.

\textbf{Architecture Overview} The Multi-Agent Spatial Handshaking \textbf{MASH} network guides the exchange of spatially-structured image data across several agents.  The general layout of the network consists of $N$ identical modules that are deployed across $N$ separate agents.  Each module consists of (1) a semantic segmentation backbone, (2) a pair of \textit{query} and \textit{key} encoders, (3) a dense correspondence calculation, (4) a context-based smoothing network, (5) an image sampling operation, and (6) an output fusion operation.  Each of these $N$ modules is processed \textit{locally} per robot, allowing for distributed computation across the robot group.

\textbf{Inputs and Outputs} The input of \textbf{MASH} consists of $N$ RGB images ($H \times W \times 3$) captured from the viewpoints of $N$ independent agents.  For sake of explanation, we denote one of these agents as the \textbf{target} agent and the others as \textbf{supporting} agents.  The final output of \textbf{MASH} is an obstruction-free semantic segmentation for the \textbf{target} agent, where ``obstruction'' refers to an unexpected (\textit{i.e.} out-of-distribution) foreground object.
Note that all agents may \textit{simultaneously} be a \textbf{target} and \textbf{supporting} agent, with shared computation and communication for each role.  Without loss of generality, we describe our method for one \textbf{target} agent.

\textbf{Semantic Segmentation Backbone} We use SegNet~\cite{badrinarayanan2017segnet} as the backbone of our spatial handshaking network.  This network architecture uses a convolutional encoder-decoder to transform a RGB input ($H \times W \times 3$) into a full-resolution semantic segmentation mask ($H \times W \times C$).  For our purposes, we do not alter the structure of the backbone network; rather, we extract and forward two of its layers to the rest of \textbf{MASH}.  Specifically, we use (A) an intermediate spatial feature map and (B) the output semantic segmentation distribution.  We note that our SegNet backbone is interchangeable with most semantic segmentation architectures of a similar structure.

\textbf{Data Compression} After extracting the intermediate feature map from the segmentation backbone, we use a pair of encoders to further process and compress it into a \textbf{query} and \textbf{key} feature map.  These two encoders consist of a sequence of learned convolutional kernels of spatial size $1 \times 1$ which transform the original spatial feature map $f$ ($H_s \times W_s \times F$) into channel-reduced maps $q$ ($H_s \times W_s \times Q$) and $k$ ($H_s \times W_s \times K$), respectively.  Following this compression, the target agent transmits $q$, and the supporting agent (or agents) receives $q$ and compares it against its own $k$.

This compression step reduces the bandwidth required for transmitting the feature map $q$, and it reduces the complexity of computing dense correspondences between the two feature maps, $q$ and $k$.  Furthermore, by having separate \textbf{query} and \textbf{key} encoders, the agents have a mechanism for describing the input image in the form of a question (\textbf{query}) and an answer (\textbf{key}), similar to the mechanism explored by When2Com~\cite{liu2020when2com}.  Previously shown for 1D feature vectors, \textit{asymmetric} encoders allow the transmitted \textbf{query} to be much smaller than the \textbf{key}, even for spatial data and dense correspondence settings.

\textbf{Data Correspondence} The spatial feature maps ($q$ and $k$) from the preceding step have asymmetrical channel dimensions ($Q$ and $K$).  
To compare these two quantities, we first convolve $q$ with $K$ learned convolutional kernels of size $1 \times 1 \times Q$ to produce a spatial feature map $p$.  
Next, using the $p$ map from the target robot (subscript $T$) and the $k$ map from the supporting robot (subscript $S$), we compute a dense pairwise distance volume:
\begin{equation}\label{eq:similarity_volume}
   \mathcal{D}_{TS}[x,y,x',y'] = -d(p_T[x,y],k_S[x',y']),
\end{equation}
where $d$ defines a distance metric (e.g. $L_2$) between two feature vectors of dimension $K$.  
Next, we reshape this 4D similarity volume into a 3D tensor of size $H_T \times W_T \times(H_S \times W_S)$.  
This new representation describes how each compressed \textbf{query} feature from the target robot $A$ matches to all \textbf{key} features from the supporting robot $B$.  
Additionally, we compute a ``no-match'' score:
\begin{equation}\label{eq:similarity_volume}
   \mathcal{D}_{T\O}[x,y] = -d(p_T[x,y],\vec{0}).
\end{equation}
We append this ``no-match'' tensor ($H_T \times W_T$) to the 3D similarity tensor, yielding a final volume of $H_T \times W_T \times(H_S \times W_S + 1)$.  
This added score gives the target robot a \textit{self-attention} mechanism for evaluating the descriptiveness of its own features, as used in prior work~\cite{avraham2019empnet, liu2020when2com}.
Finally, we convert these per-cell matching scores into a \textit{matching distribution} by applying the Softmax operation across the channel dimension.

To summarize the preceding three steps, first, the SegNet backbone encodes a low-resolution, high-dimensional spatial representation of the original image.  Next, the data compression step reduces the channel dimension of this representation for more efficient communication and computation.  We then exchange these compressed representations across agents.  Finally, the data association step compares pairs of spatial feature maps to produce a dense correspondence volume.  

\begin{figure}
\vspace{5mm}
\centering
\includegraphics[width=\linewidth]{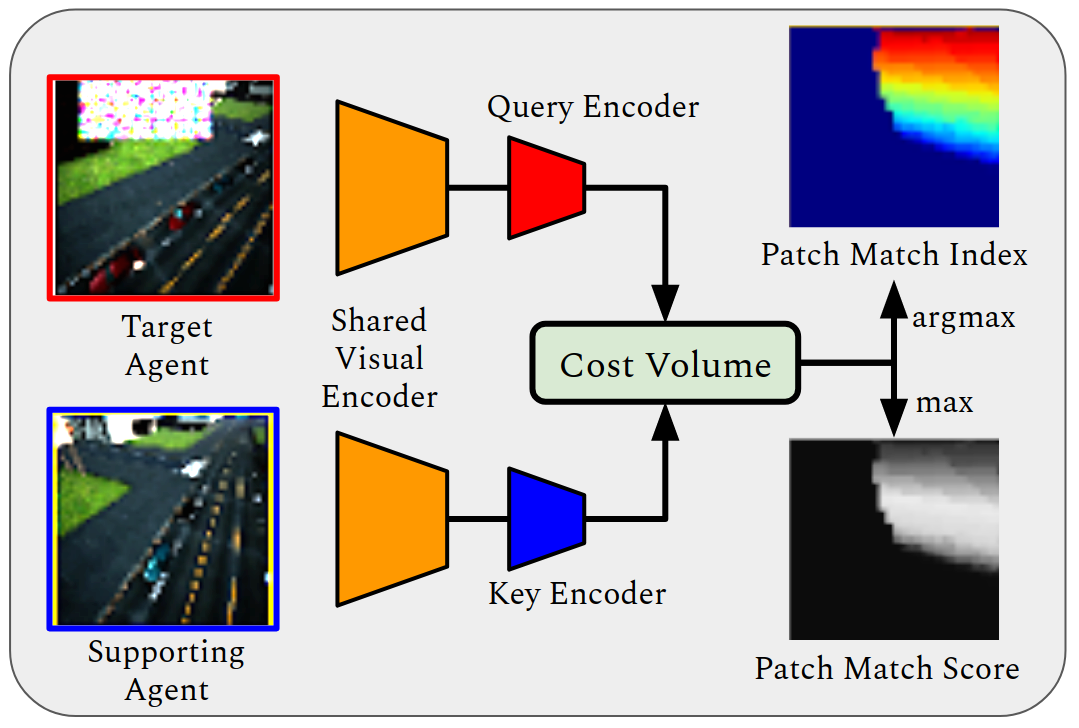}
\setlength{\belowcaptionskip}{-15pt}
\caption{\textbf{Simplified cost volume pipeline.} The target and supporting agents encode their visual inputs into query and key spatial feature maps, respectively.  We compute a smoothed cost volume from these spatial maps.  The channel-wise \textbf{argmax} of this volume describes how each source pixel maps to the target; the channel-wise \textbf{max} describes the strength of this mapping.}
\label{fig:CostVolumeSummary}%
\end{figure} 

The following steps outline how we identify and exchange image patches for multi-agent fusion:

\textbf{Correspondence Smoothing and Infilling}  The correspondence volume produced by the previous step highlights visually similar patches between pairs of agents.  To generate this volume, each patch from the target agent is matched \textit{independently} to all of the patches from the source agent; moreover, the spatial location of the target patch and the matching distributions of its neighboring patches do not factor into generating the correspondence volume.  In essence, the raw correspondence volume does not directly leverage strong prior knowledge that relates to the spatial arrangement of image patches, such as matching continuity and dataset priors.  If a given target patch $X_T$ matches to a particular source patch $Y_S$, we would expect patches in the neighborhood of $X_T$ to match to patches roughly in the neighborhood of $Y_S$.  Furthermore, for instance, if agents always capture gravity-oriented images, that should serve as a strong prior for corresponding unseen image matches.

To this end, we propose a convolutional autoencoder that smooths and infills noisy correspondence volumes.  This module takes a raw correspondence volume as input ($H_T \times W_T \times(H_S \times W_S + 1)$) and produces an identically sized correspondence volume as output.  Our convolutional architecture exploits both the spatial arrangement of the cost volume ($H_T \times W_T$) and its corresponding matching distribution ($H_S \times W_S + 1$).  We train this encoder to match the ground truth warping map using a cross entropy classification objective.

\begin{figure}
\vspace{5mm}
\centering
\includegraphics[width=\linewidth]{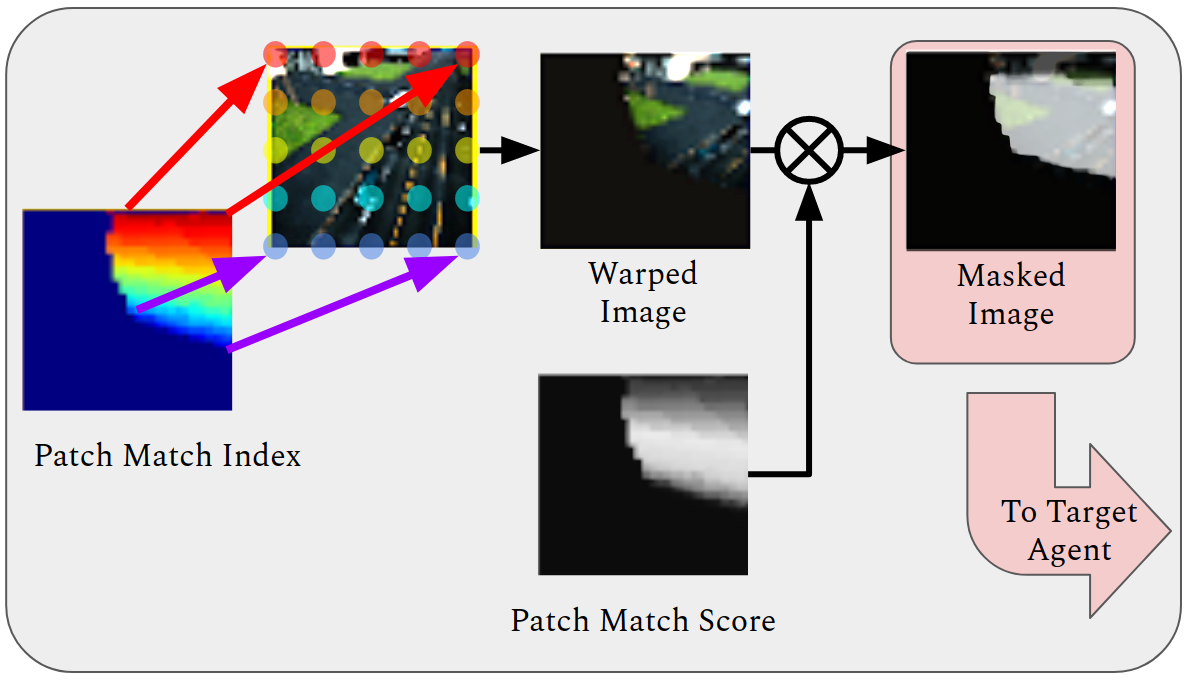}
\setlength{\belowcaptionskip}{-15pt}
\caption{\textbf{Warping pipeline.} The patch match index is used to warp patches from the supporting frame into the target frame.  Prior to re-transmission back to the target agent, the patch match confidence is used to mask (or weight) the warped data. Here we visualize the pipeline using RGB images, but the same technique applies to any image-structured data, such as segmentation masks.}
\label{fig:WarpingSummary}%
\end{figure} 

\textbf{Output Sampling and Fusion} We use the smoothed correspondence volume ($H_T \times W_T \times(H_S \times W_S + 1)$) to directly sample and weight the output contributions of the supporting robot(s).  Specifically, we compute the channel-wise \textbf{ArgMax} and \textbf{Max} of the correspondence volume to produce a 2D warping grid and 2D warping confidence map, as highlighted in Fig~\ref{fig:CostVolumeSummary}.  The warping grid describes how each spatial cell from a target agent $T$ could be complemented by a corresponding cell in a source agent $S$.  If the \textbf{ArgMax} selects the last ``no-match'' channel, then the original spatial cell from $T$ is used as the complementary cell.  The 2D warping confidence map describes the confidence associated with that best match.

As shown in Fig~\ref{fig:WarpingSummary}, we use the 2D warping grid to align visual information from the supporting agent(s) into the target frame.  In our case, we use the warping grid to align the pre-ArgMax semantic segmentation output distribution from the supporting agent(s).  

Next, we use the 2D match confidence map to determine how to best combine the semantic segmentation distributions from each agent (including itself).
Namely, we want image patches that have a high confidence of being warped properly (\textit{i.e.} those patches with a large confidence map score) to more prominently contribute to the final output.  We consider several variants that fuse the segmentation contributions from each agent:

\textit{HardSelection}: For each spatial cell, hard selection uses the semantic segmentation distribution from the agent with the strongest warping confidence score.

\textit{SoftSelection}: For each spatial cell, soft selection scales the distribution of each agent based on its warping confidence score.  The scaled semantic segmentation distributions are then fused together via addition.

\textit{StackedFusion}: Stacked fusion concatenates the warped segmentation masks from each output.  Then, it passes the resulting stack to a small convolutional network that generates the final fused segmentation mask.

Finally, after the semantic segmentation output distributions are fused, we generate a semantic mask prediction via channel-wise \textbf{ArgMax}.  

In summary, we propose a neural architecture that learns both compression \textit{and} warping in an end-to-end manner, allowing us to directly train towards the task of semantic segmentation.  Moreover, our design decisions improve resilience to obstructions (via the learned costmap autoencoder) and allow us to efficiently deploy across any number of agents in a bandwidth-limited collaborative swarm.

\textbf{Training} We require two sets of ground truth signals to train \textbf{MASH}: (1) per-agent semantic segmentation masks and (2) patch correspondences between the target and supporting agents.  We use a cross entropy loss for both semantic segmentation and patch correspondence.  Additionally, though the network is trained in a dense, centralized manner, it can be deployed in a distributed manner.  During inference, each agent receives the same trained network weights; and unlike centralized models, full network connectivity is not a hard requirement.  We show results for optional top-1 communication (where we communicate the feature with the best warping score, referred to as \textit{HardSelection}).  We emphasize that, during inference, the only inputs to the network are a single RGB image per robot.

\textbf{Bandwidth and Computation} We designed the \textbf{MASH} architecture to reduce communication bandwidth while leveraging a swarm's access to distributed computing.  Regarding communication bandwidth, we reduce the naive communication of all raw images ($(N-1) \times H \times W \times 3$) to:

\begin{equation}\label{eq:bandwidth}
\begin{split}
    \overbrace{
        \underbrace{(H_s \times W_s \times Q)}_\text{Query Map}
    }^\text{Target Agent (Tx)} +
    \underbrace{(N-1)}_\text{Agents} \times
    \overbrace{
        \underbrace{(H \times W \times C)}_\text{Segmentation Map}
        \times
        \underbrace{r}_\text{Selection Factor}
    }^\text{Supporting Agent (Rx)}
\end{split}
\end{equation}

\noindent with $H$ and $W$ image dimensions, $C$ classes, $H_s$ and $W_s$ downsampled spatial dimensions, $Q$ compressed channel dimension ($Q \leq K$), $N$ agents, and $r$ selection factor ($0 \leq r \leq 1$).  

Note, the selection factor $r$ denotes the \textit{empirical} fraction of spatial information transmitted by the supporting agent, and it reflects the degree of overlap between pairs of agents.  This factor ranges from $0$ (for pairs with no overlap) to $1$ (for pairs with full overlap); it is generally quite small ($r << 1$), especially for our dataset that includes intermittently and partially overlapping agents.

Regarding computation, rather than perform processing operations in a centralized manner, we distribute and parallelize several operations across the available agents, reducing both latency and computational burden.  Please refer to Fig~\ref{fig:overall} to see how operations are grouped and parallelized.
\begin{table*}
\vspace{5mm}
\resizebox{\linewidth}{!}{%
\begin{tabular}{llc|c||c|c||cc||cc||cc}
\\
& & \multicolumn{4}{c}{Segmentation} & \multicolumn{2}{c}{MASH Parameters} & \multicolumn{2}{c}{Bandwidth} & \multicolumn{2}{c}{Compute Time} \\
\midrule
& & \multicolumn{2}{c}{\textit{\textbf{Full}} Image} & \multicolumn{2}{c}{\textit{\textbf{Obstructed}} Region} & Query & Key & Per & Per & Per & Per  \\
& & Accuracy & IoU & Accuracy & IoU & Dim & Dim & Target & Supporting & Target & Supporting \\
\midrule
Baselines & Inpainting~\cite{cai2017blind} (Trained on Target View) & 50.29 & 40.70 & 27.54 & 18.66 & - & - & - & - & 1.00$t$ & - \\ 
& Inpainting~\cite{cai2017blind} (Trained on All Views) & 55.24 & 44.77 & 22.85 & 15.01 & - & - & - & - & 1.00$t$ & - \\
& InputStack~\cite{detone2016deep,wang2017deepvo} & 34.85 & 26.88 & 23.35 & 15.83 & - & - & - & 1.00$x$ & 1.00$t$ & - \\
& OutputStack & 51.35 & 40.94 & 27.65 & 17.21 & - & - & - & 1.00$y$ & 1.08$t$ & 1.00$t$ \\
& WarpedOutputStack*~\cite{liu2020who2com} & \textbf{64.16} & \textbf{54.55} & \textbf{38.31} & \textbf{28.27} & - & - & - & - & - & - \\
\midrule
MASH & vs Bandwidth Consumption & 57.95 & 43.29 & 26.16 & 16.63 & 16 & 64 & 0.08$x$ & 0.09$y$ & 1.09$t$ & 1.69$t$ \\
& & 57.34 & 45.68 & 25.70 & 16.86 & 32 & 64 & 0.17$x$ & 0.08$y$ & 1.09$t$ & 1.70$t$ \\
& & 65.15 & 55.70 & \textbf{33.59} & \textbf{23.73} & 64 & 64 & 0.33$x$ & 0.04$y$ & 1.09$t$ & 1.69$t$ \\
& & 64.52 & 53.76 & 31.04 & 21.63 & 64 & 128 & 0.33$x$ & 0.03$y$ & 1.09$t$ & 1.71$t$ \\
& & \textbf{65.60} & \textbf{56.30} & 32.45 & 22.00 & 64 & 256 & 0.33$x$ & 0.03$y$ & 1.09$t$ & 1.68$t$ \\
\bottomrule
\end{tabular}}
\setlength{\belowcaptionskip}{-15pt}
\caption{\textbf{Segmentation Results for Full Image and Obstructed Region.}
We compare baselines and \textbf{MASH} variants using metrics such as bandwidth consumption, compute time, mean accuracy, and mean IoU.  
Here, the bandwidth used by the target agent is based on raw input size ($x = 256 \times 256 \times 3$); the bandwidth used by each supporting agent is based on output semantic mask size ($y = 256 \times 256$); and the compute time is based on SegNet's compute time ($t \approx 0.2s$ on an Nvidia RTX 2080 GPU), averaged over 50 inference runs.  The transmitted bandwidth is determined by the channel dimension of the compression encoders, and the received bandwidth is empirically calculated based on the matched fraction of the image.  *This baseline uses the auxiliary ground truth warping grid (\textit{i.e.} extra ground truth information) and is an upper bound.  
}
\label{tab:mash_table_reorder}
\end{table*}

\section{EXPERIMENTS}
\textbf{Dataset}. 
Our dataset consists of synchronized RGB and semantic segmentation images captured from 6 moving agents in a simulated photo-realistic urban environment.  The fields of view of these agents periodically and non-trivially overlap, and their cameras capture both static and dynamic scene objects.  Additionally, to provide an intermediate training signal, our dataset also includes ground truth dense pixel correspondences between pairs of images, computed from the raw pose and depth provided by the simulator.  

The task associated with this dataset is the following: using RGB data captured from single snapshot in time (\textit{i.e.} 6 synchronized images from different parts of the swarm), the target agent must generate the most accurate semantic segmentation mask.  Additionally, we constrain this task by restricting the direct exchange of raw RGB data; namely, the target agent may access its own RGB data at no cost, but communication with other agents is penalized based on bandwidth consumption.  Furthermore, to better tailor our dataset to investigating the task of \textit{perceptual resilience} in a swarm, we spawn random obstructions in the 3D scene (such as birds and trees).  Each target agent must then predict an unobstructed semantic segmentation mask, despite the presence of a foreground obstruction.

To produce this dataset, we spawned a swarm of 6 drones in the photo-realistic, multi-robot AirSim~\cite{airsim2017fsr} simulator.  We commanded the drones to move roughly together throughout the environment, capturing synchronized images along the way.  
To ensure dynamically overlapping views, we commanded different drones to travel with different yaw rates.  
Additionally, each time the swarm arrived at an intersection, we commanded each drone to randomly and independently explore the intersection, further ensuring interesting frame overlaps.  
We repeated this process until the swarm explored all of the road intersections in the simulator.

Futhermore, we synthesize an \textit{obstruction-prone} variant.  We post-processed each snapshot of the base AirSim dataset, randomly placing a 3D object into the scene and rendering this object in the views of each observer.  We used the pose and depth information from each agent and the pose and depth information of the 3D obstruction model to perform the insertion rendering.

In summary, our dataset consists of $18,588$ snapshots across $6$ non-static robots.  The robots have an average pose difference of $8.7 m$ and $50^{\circ}$.  Sporadic motion and yaw of the robots creates intermittently and partially overlapping views.  A random pair of agents has an average pixel overlap of $37\%$ with a standard deviation of $15\%$.  For our experiments, we select one of the $6$ agents as the \textit{target} and allow it to communicate with any number of the $5$ \textit{supporting} agents.

\textbf{Baselines.} We compare against several baselines:

\textit{Inpainting} uses the SegNet architecture to directly transform a (potentially obstructed) target image into an unobstructed semantic segmentation mask, as inspired by~\cite{cai2017blind}.  The model is trained to ``hallucinate'' a reasonable prediction for obstructed pixels, rather than query the other frames that have a glimpse at those occluded pixels.

\textit{InputStack} concatenates the unwarped RGB images from all agents, and it forwards this stack of images to SegNet for semantic segmentation.  This procedure resembles that of Deep Homography and Visual Odometry methods~\cite{detone2016deep,wang2017deepvo}.

\textit{OutputStack} uses a shared SegNet model to process each unwarped RGB input into a semantic segmentation output distribution.  Next, it concatenates these output distributions and then fuses them through a learned convolutional kernel.  

\textit{WarpedOutputStack} is identical to \textit{OutputStack} with the exception that the output distributions are warped to the target frame using the ground truth warping grid.  This warping grid is normally provided during training only, so this baseline approximates the upper-bound performance of a network which learns that warping.

\textbf{Results.}\label{results}
As shown in Table~\ref{tab:mash_table_reorder}, \textbf{MASH} achieves superior performance on segmentation metrics, while simultaneously improving bandwidth consumption and computational parallelism.  For ease of comparison, we show bandwidth and compute time in terms of fixed quantities, such as input image size ($x$), output image size ($y$), and SegNet compute time ($t$).  We also show segmentation accuracy and mean IoU evaluated both on the full image and within its obstructed pixels.  Please refer to Fig~\ref{fig:Qualitative_Infill} for a reference on how the obstructed region is scored--for this metric, the mean IoU score is only computed on the pixels within the inserted obstruction (\textit{e.g.} the unknown teal eagle class in Fig~\ref{fig:Qualitative_Infill}).

\textit{Segmentation Performance}. We observe a couple of interesting trends.  Despite not having access to unobstructed data, the \textbf{Inpainting} baselines produce reasonable semantic masks, even in the obstructed region of the image.  This trend likely results from their ability to ``hallucinate'' plausible outputs from the surrounding input context.  The tested dataset also has a strong scene prior for static classes, further enhancing the effectiveness of inpainting methods.  The \textbf{InputStack} and \textbf{OutputStack} baselines have the opposite problem to \textbf{Inpainting}; even though they can access unobstructed data, their simple, stacked architectures do not explicitly correspond complementary image features, and therefore  the extra information detracts from overall performance.  The importance of image correspondence is demonstrated by the strong performance of the \textbf{WarpedOutputStack} upper bound, which uses the otherwise inaccessible ground truth warping information to align the output feature masks.

In the bottom of Table~\ref{tab:mash_table_reorder}, we present the results for our \textbf{MASH} network.  Of our three proposed multi-agent fusion mechanisms, we highlight the results for the \textbf{HardSelection} mechanism, which yielded the best trade-off between segmentation performance and bandwidth consumption.  Each row shows the performance of this variant under various bandwidth constraints.  Our \textbf{MASH} models achieve a $~11\%$ improvement in \textit{\textbf{Full} Image} accuracy and mean IoU over several key baselines.  We similarly observe a $~6\%$ improvement in \textit{\textbf{Obstructed} Region} accuracy and mean IoU.  

These significant performance gains support several of our design choices.
We emphasize that \textbf{MASH} does not fall prey to the shortcomings of the baselines: our network is able to benefit from multiple, unaligned sources of image data via a learned, data-efficient warping mechanism.  Furthermore, our novel cost volume autoencoder enables obstructed image regions to be properly corresponded across multiple viewpoints, resulting in improved segmentation of degraded regions.  

Additionally, we demonstrate comparable performance of the strong \textbf{WarpedOutputStack} upper bound.  We attribute this narrow gap to the ability of \textbf{MASH} to properly align contributing data as well as its ability to intelligently select between these contributions.  Both of these features are present within the baseline, but the \textbf{MASH} network has a slightly deeper network depth which yields slightly stronger inference capabilities.

\textit{Bandwidth}. In the bottom of Table~\ref{tab:mash_table_reorder}, we show that we are able to tune bandwidth consumption and segmentation performance by manipulating the channel dimension of the transmitted \textit{query}, as shown by the \textit{Bandwidth Per Target} column.  As we relax our bandwidth constraint (by increasing the query dimension), we observe an improvement in perception performance.  This improvement matches expectations, especially since each agent can then pack more information into its query map.  

By altering the query dimension, we can dictate how a \textit{target} agent consumes bandwidth.  However, \textbf{MASH} itself moderates how each \textit{supporting} agent consumes bandwidth.  Namely, the warping score produced by \textbf{MASH} guides how much output information is selectively transmitted from the supporting agents to the target agent.  The \textit{Bandwidth Per Supporting} column shows the amount of data (on average across the dataset) that \textbf{MASH} passes from each supporting agent to the target agent.  Our dataset consists of largely non-overlapping data sources, and \textbf{MASH} is able to selectively transmit data that improves its perception performance, while avoiding a substantial fraction of distractor data.

\textit{Compute}. We show that \textbf{MASH} yields improved performance (comparable to a ground truth augmented variant, \textbf{WarpedOutputStack}) at a reasonable, \textit{distributed} computational overhead.  The target agent only introduces an additional $9\%$ of compute time in addition to a backbone SegNet architecture.  And each supporting agent only incurs a $70\%$ increase in compute time.  Note that rather than burdening the target agent with each per-agent correspondence, warping, and fusion module, these computations are effectively distributed among the supporting agents.
\section{CONCLUSION AND FUTURE WORK}
We propose a variant on the multi-agent collaborative perception task in which robots must communicate visual information to improve their local perception performance.  We adapt this general task to target \textit{bandwidth-limited} and \textit{obstruction-prone} semantic segmentation for networked vehicles.  To identify and exchange relevant information in this restricted multi-agent setting, we introduce a neural network architecture that uses learned (1) spatial handshake communication, (2) cost volume correspondence, (3) context-aware cost volume smoothing, and (4) attention-based output fusion.  Furthermore, we show that our model is able match the performance of warping-based upper bound, with reduced bandwidth and computational overhead.
Future work in this domain will explore sparse image correspondence methods to further reduce transmission of unnecessary matching data between agents.  Our work primarily addresses the task of multi-agent perception without active control; however, future extensions will involve active multi-agent perception and in-the-loop control.

\section{Acknowledgement}
\label{sec:acknowledgement}
\noindent This work was supported by ONR grant N00014-18-1-2829.

\printbibliography
\end{document}